\documentclass[final, twocolumn]{elsarticle}

\journal{arXiv}

\usepackage{amssymb}  
\usepackage{amsthm}
\usepackage{amsmath}
\usepackage{booktabs}
\usepackage{enumitem}
\usepackage{graphicx}
\usepackage{subfig}
\usepackage{siunitx}
\usepackage[table]{xcolor}

\begin{document}

\begin{frontmatter}

\title{Automated detection of corrosion in used nuclear fuel dry storage canisters using residual
neural networks}

\author[address01]{Theodore Papamarkou}
\author[address02]{Hayley Guy}
\author[address03]{Bryce Kroencke}
\author[address04]{Jordan Miller}
\author[address05]{Preston Robinette}
\author[address06]{Daniel Schultz}
\author[address01]{Jacob Hinkle}
\author[address07]{Laura Pullum}
\author[address07]{Catherine Schuman}
\author[address08]{Jeremy Renshaw}
\author[address09]{Stylianos Chatzidakis}

\address[address01]{Computational Sciences and Engineering Division, Oak Ridge National Laboratory, Oak Ridge, Tennessee, 
USA}
\address[address02]{Department of Mathematics, North Carolina State University, Raleigh, North Carolina, USA}
\address[address03]{Department of Computer Science, University of California, Davis, California, USA}
\address[address04]{Center for Cognitive Ubiquitous Computing, Arizona State University, Tempe, Arizona, USA}
\address[address05]{Presbyterian College, Clinton, South Carolina, USA}
\address[address06]{Innovative Computing Laboratory, University of Tennessee, Knoxville, Tennessee, USA}
\address[address07]{Computer Science and Mathematics Division, Oak Ridge National Laboratory, Oak Ridge, Tennessee, USA}
\address[address08]{Electric Power Research Institute, Palo Alto, California, USA}
\address[address09]{Reactor and Nuclear Systems Division, Oak Ridge National Laboratory, Oak Ridge, Tennessee, USA}





\begin{abstract}
Nondestructive evaluation methods play an important role in ensuring component integrity and safety in many industries. 
Operator fatigue can play a critical role in the reliability of such methods. This is important for inspecting high value assets or 
assets with a high consequence of failure, such as aerospace and nuclear components. Recent advances in convolution neural 
networks can support and automate these inspection efforts. This paper proposes using residual neural networks (ResNets) for 
real-time detection of corrosion, including iron oxide discoloration, pitting and stress corrosion cracking, in dry storage stainless 
steel canisters housing used nuclear fuel. The proposed approach crops nuclear canister images into smaller tiles, trains a 
ResNet on these tiles, and classifies images as corroded or intact using the per-image count of tiles predicted as corroded by the 
ResNet. The results demonstrate that such a deep learning approach allows to detect the locus of corrosion via smaller tiles, and 
at the same time to infer with high accuracy whether an image comes from a corroded canister. Thereby, the proposed approach 
holds promise to automate and speed up nuclear fuel canister inspections, to minimize inspection costs, and to partially replace 
human-conducted onsite inspections, thus reducing radiation doses to personnel.
\end{abstract}


\end{frontmatter}


\section{Introduction}

\begin{figure*}[t!]
	\centering
	\includegraphics[scale=0.25]{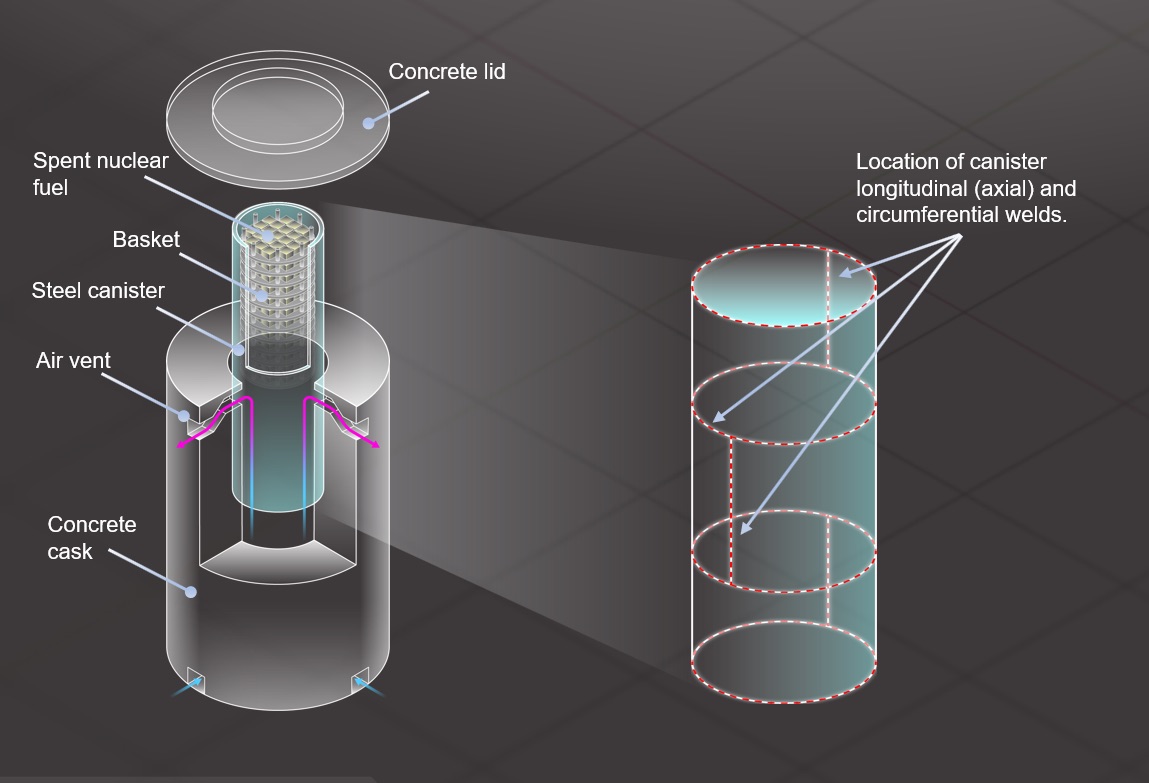}
	\caption{Illustration of a vertical used nuclear fuel dry storage system.
	This figure is for illustration purposes and does not attempt to depict a specific design.
	There is a large variation in design characteristics, which is not possible to capture in a figure, e.g.,
	dry storage systems that use a steel overpack for shielding instead of a concrete one
	or stainless steel canisters made by welding two cylindrical sections instead of three.}
	\label{fig:storage}
\end{figure*}

Dry storage systems housing used nuclear fuel from commercial nuclear power reactors (Figure \ref{fig:storage}) will be used for 
longer periods than initially anticipated and there is a concern that some of these systems may become vulnerable to physical 
degradation, e.g., pitting and chloride-induced stress corrosion cracking (SCC). The potential for SCC at the heat affected zones 
of welded stainless-steel interim storage canisters has been identified by several agencies
\cite{kusnick2013finite,rigby2010evaluation,hanson2012gap,chopra2014managing,chu2014flaw,Gorman2014,FuhrK.2017,
Broussard2015}.
Although this concern has 
motivated the development of delivery systems and inspection of these canisters on a regular basis
\cite{Meyer2016,Meyer2013,Renshaw2016b,Chu2019,BryanC.2016,Tang2019,ChatzidakisStylianos2018,StylianosChat2019,
Chatzidakis2020,Chatzidakis2019,Chatzidakis2018,Chatzidakis2017,Lin2019,Renshaw2016a,Renshaw2015,RenshawJ2019,
ChuShannonand2019},
the large number of 
stored canisters that need to be visually inspected (currently approximately $3,200$ canisters have been installed in the U.S. and 
projections show that eventually over $10,000$ canisters will be required by 2050
\cite{JonesJr.andRobert2015}), high radiation levels, limited access 
(through small size vents) and space constraints (overpack-canister gap less than 5-10 centimeters) make in-situ visual 
inspections challenging. This necessitates the development of remotely operated systems for real-time detection of flaws 
including type, location, size, and density, to support future remediation activities, reduce operator errors, and optimize repair 
quality. As a result, an integrated deep learning framework that would allow scanning a larger area and with high detection 
capabilities would be a big step forward to reduce dose-rate to operators, minimize inspection costs, and ensure long-term 
safety of used nuclear fuel canisters.

Existing methods for detecting structural defects in images include Frangi filters and Hessian affine region detectors
\cite{yeum2015vision}, total 
variation denoising
\cite{rudin1992nonlinear,cha2016vision},
combinations of image processing techniques for feature extraction
\cite{o2014regionally,wu2016improvement}
with machine learning 
algorithms for classification
\cite{lecun1998gradient,butcher2014defect},
and more lately deep convolutional neural networks (CNNs)
\cite{soukup2014convolutional,cha2017deep}.
So far, no deep 
learning approaches have been used to address inspection of used nuclear fuel canisters. Radiation levels typically create 
“noise”, “flacking”, “snowing” and other artifacts in cameras. Such artifacts pose a challenge for conventional feature detection 
techniques, e.g., shallow neural networks, that rely on clean surfaces, isolated defects, and radiation-free viewing with optimal 
lighting conditions. So, the process of inspecting used nuclear fuel canisters is not automated algorithmically and it induces a 
challenging image classification problem. Deep learning is a plausible approach for automating such a challenging classification 
problem. In fact, 
\cite{cha2017deep}
have applied CNNs for detecting concrete cracks in radiation-free civil infrastructures. Along the lines of
\cite{cha2017deep},
this paper crops images to smaller tiles to train the underlying CNN.

The proposed approach is differentiated from
\cite{cha2017deep}
in three ways. Firstly, ResNets are used for detecting corrosion instead of 
deploying a custom CNN architecture. Training three different ResNets (ResNets-18, ResNet-34 and ResNet-50) demonstrates 
that the proposed approach tackles the corrosion detection problem without being sensitive to the choice of ResNet 
architecture.
Secondly, the present paper introduces a classification rule for identifying images as corroded or intact using ResNets trained on 
tiles cropped out of these images. This way, the locations of potential corrosion are linked to tiles predicted as corroded, while at 
the same time the problem of classifying the original images is solved with high prediction accuracy without being sensitive to 
tile-specific prediction errors. Finally, the proposed approach is focused on the specific application of detecting
corrosion in real data imagery similar to used nuclear fuel canisters.

CNNs have been applied for detecting cracks in videos of nuclear power plants \cite{chen2018}.
The proposed approach differs from \cite{chen2018} in four aspects.
Firstly, data have been collected from nuclear fuel canisters, and not from nuclear power plants.
Secondly, the EPRI data consist of images (spatial information), and not of videos (spatiotemporal information).
Thirdly, the classification problem in this paper is to decide whether each image contains any sign of corrosion,
whereas the classification problem in \cite{chen2018} is to decide whether tubelets extracted from multiple video frames
contain corrosion cracks.
Fourthly, this paper employs ResNets for tile classification in conjunction with a custom image classification rule,
whreas \cite{chen2018} employs a custom CNN architecture for crack patch detection per frame
along with a three-step data fusion algorithm for generating bounding boxes around detected crack tubelets.

\section{Materials and methods}

\subsection{Data}

\begin{figure*}[t!]
\parbox{.48\linewidth}{
\centering
\vspace{-0.12in}
\includegraphics[scale=0.275]{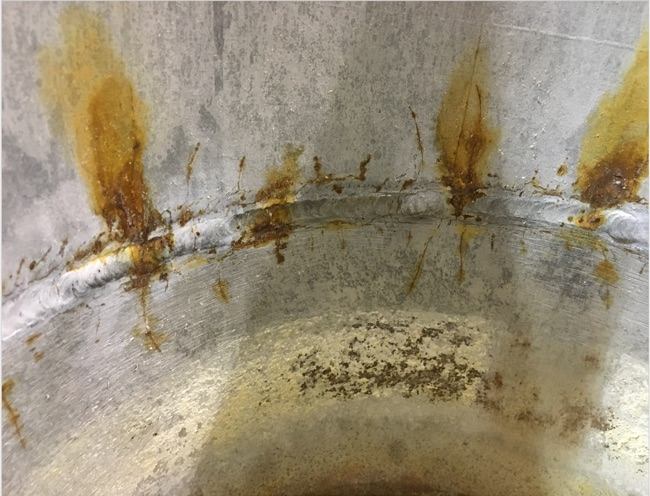}
\caption{Example image with visible corrosion, stress corrosion cracks, and artifacts (scratches, shadows, etc.). Note: this is a 
	laboratory corroded stainless steel sample and not an image from an actual used fuel canister; to the best of our knowledge no 
	canisters have been detected with defects).}
\label{fig:example_image}
}
\hfill
\parbox{.48\linewidth}{
\subfloat[Corroded]{{\includegraphics[scale=0.19]{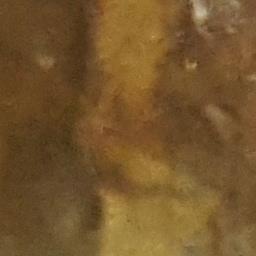}}\label{fig:3aa}}\
\subfloat[Corroded]{{\includegraphics[scale=0.19]{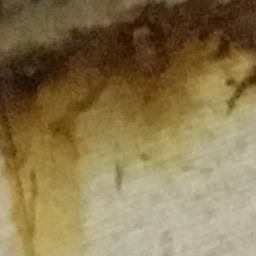}}\label{fig:3ab}}\qquad
\subfloat[Intact]{{\includegraphics[scale=0.19]{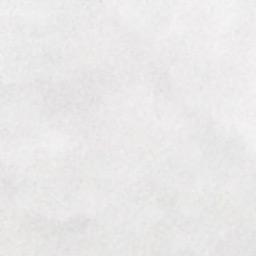}}\label{fig:3ba}}\
\subfloat[Intact]{{\includegraphics[scale=0.19]{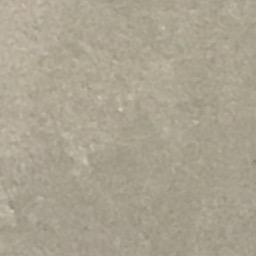}}\label{fig:3bb}}\\
\subfloat[Scratched]{{\includegraphics[scale=0.19]{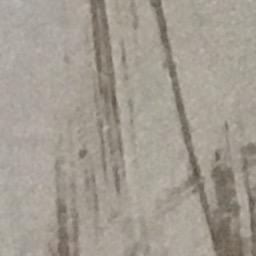}}\label{fig:3ca}}\
\subfloat[Scratched]{{\includegraphics[scale=0.19]{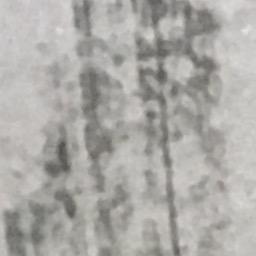}}\label{fig:3cb}}\qquad
\subfloat[Shadows]{{\includegraphics[scale=0.19]{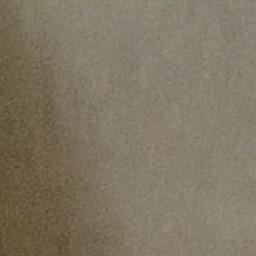}}\label{fig:3da}}\
\subfloat[Shadows]{{\includegraphics[scale=0.19]{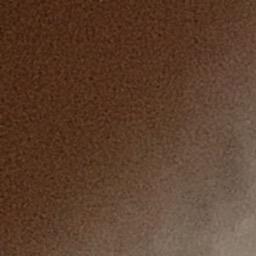}}\label{fig:3db}}
\caption{Examples of $256\times 256$-pixel resolution tiles used for training, which demonstrate the large variability in the 
	dataset. Tiles with
	with corrosion, including iron oxide discoloration, pitting or stress corrosion cracks (with or without iron oxide discoloration)
	(\ref{fig:3aa}-\ref{fig:3ab}) are labeled as corroded, while tiles without 
	defects (\ref{fig:3ba}-\ref{fig:3bb}) or with scratches (\ref{fig:3ca}-\ref{fig:3cb}) or shadows (\ref{fig:3da}-\ref{fig:3db}) are 
	labeled as intact.}%
\label{fig:example_corrosion}
}
\end{figure*}

The research presented in this paper is based on data provided by the Electric Power Research Institute (EPRI). The data consist 
of 166 images taken from flaw mockup specimens produced by EPRI. These images, which capture stainless-steel canister 
surfaces, were taken with a $16$-megapixel camera from a variety of locations, angles, and lighting conditions to induce high 
variability in the dataset. Each image has a $4,928\times 3,264$-pixel resolution. Figure \ref{fig:example_image} shows an 
example corroded image. Interest is in detecting corrosion in the images.
The images contain stress corrosion cracks with and without iron oxide discoloration. Interest is in detecting pitting or stress 
corrosion cracks independent of iron oxide discoloration in the images.
	
Use of deep learning for detecting
nuclear canister corrosion requires training sets of sample size larger than the $166$ available images. 
For this reason, the $166$ images were cropped into smaller tiled images of $256\times 256$-pixel resolution using a 
custom-made algorithm for this task, producing a total of $37,719$ tiled images (referred to as tiles thereafter). Each tile was 
labeled either as corroded or as intact, depending on whether it contains any signs of corrosion
(including iron oxide discoloration, pitting, and stress corrosion cracks with or without iron oxide discoloration)
or not. Thus, the ground truth for image 
annotation takes into account only corrosion identifiable via the naked human eye. The generated dataset of tiles includes a 
broad range of image variations (see Figure \ref{fig:example_corrosion}), which are necessary for CNN training.

The $256\times 256$-pixel resolution for tiles was chosen empirically. Such a small resolution facilitates the training process of 
CNNs with tiled images coming from original images of varied resolution, therefore making the approach camera-independent 
and more generally applicable. On the other hand, tiles of smaller resolution may lead CNNs to mistake elongated features, such 
as scratches, for cracks. In addition, smaller tiles render their labeling less obvious to the human eye.

\begin{table*}[t!]
\centering
\subfloat[Dataset of original images]{
\begin{tabular}{l|rrr}
	\toprule
	& Corroded & Intact & Total \\ \midrule
	Training & 50 & 49 & 99 \\
	Validation & 17 & 17 & 34 \\
	Test & 17 & 16 & 33 \\
Total & 84 & 82 & 166 \\ \bottomrule
\end{tabular}
\label{table:1a}
} \qquad
\subfloat[Dataset of tiles]{
\begin{tabular}{l|rrr}
	\toprule
	& Corroded & Intact & Total \\ \midrule
	Training & 2,898 & 19,317 & 22,215 \\
	Validation & 1,235 & 6,303 & 7,538 \\
	Test & 824 & 7,142 & 7,966 \\
	Total & 4,957 & 32,762 & 37,719 \\ \bottomrule
\end{tabular}
\label{table:1b}
}
\caption{Total number of images and number of corroded and intact images per set in the original dataset of images 
(\ref{table:1a}), from which the tiles were extracted, and in the generated dataset of tiles (\ref{table:1b}). Rows represent 
datasets and columns represent classes.}
\label{table:1}
\end{table*}

$4,957$ out of the $37,719$ tiles were manually labeled as corroded, while the remaining $32,762$ were labeled as intact. After 
manually labeling each tile either as corroded or as intact, the dataset of tiles was split randomly into a training, validation, and 
test set, containing $22,215$, $7,538$, and $7,966$ tiles, respectively, thus attaining a $60\%-20\%-20\%$ split. In order to 
keep the training, validation, and test sets independent, tiles from a particular image were included in only one set. Table 
\ref{table:1a} shows the number of corroded and intact original images while Table \ref{table:1b} shows the number of corroded 
and intact tiles in each of the three sets. An original image is corroded if it contains at least one tile manually labeled as corroded, 
otherwise it is intact. It is noted that the $60\%-20\%-20\%$ split between training, validation, and test set is preserved at the 
level of original images too ($99$, $34$ and $33$ original images in the respective sets).

\subsection{Brief overview of CNNs}

The first computational model for neural networks were conceived in the 1940s
\cite{mcculloch1943}.
Hebian networks were simulated by \cite{farley1954}
and perceptrons were created by \cite{rosenblatt1958} in the 1950s.
The neocognitron, the origin of the CNN architecture, was introduced by \cite{fukushima1980} in the 1980s.
The neocognitron introduced two basic layer types in CNNs, namely convolutional layers and downsampling layers.
LeNet, a CNN developed by \cite{lecun1989, lecun1998} in the 1990s to recognize hand-written zip code numbers,
led to the emergence of CNNs and laid the foundations of modern computer vision.

\begin{figure*}[t!]
	\centering
	\includegraphics[scale=0.6]{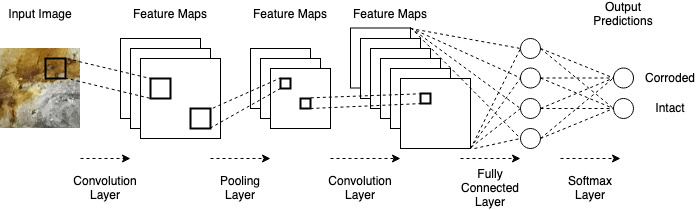}
	\caption{Visual representation of a typical feed-forward CNN architecture.}
	\label{fig:cnn_visualization}
\end{figure*}

A feed-forward CNN consists of multiple layers of units, starting with an input layer, followed by combinations of convolution, 
pooling, activation and fully connected layers, and ending with an output layer (see Figure \ref{fig:cnn_visualization}). 
Convolution layers perform convolution operations to extract high level features from the input image. Convolution operations 
preserve the spatial relationship between pixels by learning image features using small squares of input data. Each convolution 
layer is typically followed by a unit-wise activation function, such as the rectified linear unit (ReLU). Pooling layers reduce the 
dimensionality of each extracted feature, but they retain the most important information. In a fully connected layer, each neuron 
is connected to every neuron in the previous layer. Fully connected layers learn non-linear combinations of the features identified 
by the convolutional layers. A common final layer in CNNs is the softmax layer, which assigns probabilities to each class label. 
For a more detailed introduction to CNNs, see for example
\cite{goodfellow2016deep}.

CNNs are prominent tools for image classification for several reasons. In particular, CNNs
\setlist{nolistsep}
\begin{itemize}[noitemsep]
\item are universal function approximators \cite{cybenko1989},
\item attain high predictive accuracy in comparison to other machine learning algorithms,
\item automate feature extraction from input data while preserving the spatial relationship between pixels,
\item reduce the number of parameters via the convolution operator relatively to fully connected artificial neural networks, and
\item allow transfer of knowledge across different domains.
\end{itemize}

The capacity of CNNs to transfer knowledge across different domains is known as transfer learning \cite{pratt1993}.
In practice, transfer learning entails the following steps; a CNN is trained on a dataset related to some problem,
the knowledge acquired (such as the CNN weights and biases) is stored,
and the stored knowledge for this pre-trained CNN is used for fine tuning the CNN on a dataset
related to a different, yet usually relevant, problem.
In transfer learning,
some components (such as weights or biases) of a CNN are typically fixed (frozen)
to values obtained by pre-training on some dataset,
while new layers or components attached to the pre-trained CNN are trained on a new dataset.

\subsection{Proposed algorithm}
\label{prop_alg}

In this paper, ResNets are used, which are a specific type of CNN architecture
\cite{He_2016_CVPR}.
The last (output) layer of ResNets is a fully connected layer of $1,000$ units.
Since this paper deals with the binary classification problem of whether each image is corroded or intact,
the last fully connected ResNet layer of $1,000$ units is replaced by a fully connected layer of $2$ units.
The ResNet backbone preceding the replaced last layer is not frozen.
This means that the weights and biases of the ResNet backbone are not fixed to some pre-trained values,
therefore transfer learning is not employed.
Instead, the modified CNN architecture, arising from the altered last ResNet layer,
is trained by randomly initializing the weights and biases of all layers.

At a greater level of detail,
the last fully connected layer of $1,000$ units
is replaced by a
layer that concatenates adaptive average pooling and
adaptive max pooling,
a flattening layer
and a fully connected layer of $2$ units.
The implementation is based on the \texttt{fastai} library.
More details can be found in the documentation of the \texttt{cnn\_learner} method
of \texttt{fastai} and in the accompanying code of this paper (see Section \ref{tuning_and_implementation}).

\begin{figure*}[t!]
\centering
\includegraphics[scale=0.8]{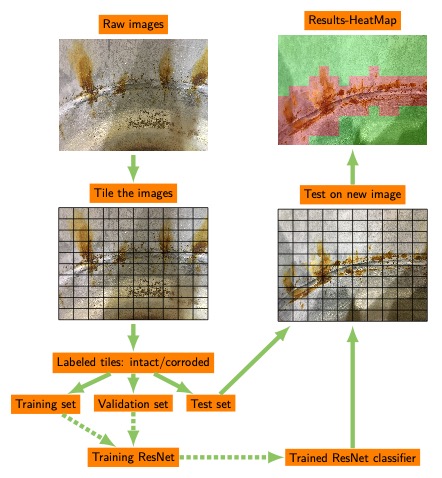}
\caption{Flowchart for classifying tiles using a ResNet. Solid lines represent training steps and dashed lines refer to test steps. 
Tiles from the test set, stitched back together to display the image of origin, are colored as red or green depending on whether 
they have been predicted as corroded or as intact by the ResNet. }
\label{fig:flowchart}
\end{figure*}

Each of the $7,719$ tile images in the 
generated dataset is passed to the ResNet input layer as a $256\times 256\times 3$ tensor, whose three dimensions correspond 
to tile height, width and RGB channel (red, green and blue), respectively. The prediction made at the ResNet output layer 
indicates whether the input tile is considered to be corroded or intact. Raw images are cropped into tiles, which are in turn 
manually labeled and split into training, validation and test set. After training the ResNet on the training set, predictions are made 
for tiles coming from the test set. The heatmap of Figure \ref{fig:flowchart} visualizes the tile predictions made by the ResNet, 
with red-colored and green-colored tiles predicted as corroded and intact, respectively.

The posed research question is to classify images as corroded or intact, whereas a ResNet provides predictions for fragments of 
the image. To solve the original problem, a rule for image classification can be defined by utilizing tile predictions. One 
classification rule is to predict the original image as corroded if it contains at least one tile predicted as corroded. However, such 
a classification rule is prone to produce false alarms due to being sensitive to tile prediction errors. Misclassifying a single tile as 
corroded is sufficient to produce a false alarm for the original image.

To make corrosion detection for an image less susceptible to its constituent misclassified tiles,
the classification of the image can be 
defined by considering the absolute frequency of misclassified tiles. Let $n_i$ be the number of tiles in the $i$-th image and
$\hat{y}_{t}^{(i)}$ the prediction made by the ResNet for tile $t$ in image $i$. The prediction $\hat{y}^{(i)}$ for image i is set to
\begin{equation}
\hat{y}^{(i)}=
\left\{
\begin{array}{ll}
1, & \mbox{if}~\displaystyle\sum_{t=1}^{n_i}\hat{y}_{t}^{(i)} > c,\\
0, & \mbox{otherwise}.\\
\end{array}
\right.
\label{eq:thres}
\end{equation}
assuming that $1$ and $0$ correspond to presence and absence of corrosion, and $c$ is a hyper-parameter.

The hyperparameter $c$ sets a threshold on the count of tiles predicted as corroded, above which the image consisting of these 
corroded tiles is predicted to be corroded itself. This classification rule has been instigated by training ResNets on tiles. It has 
been observed empirically that images
with pitting or stress corrosion cracks tend to contain higher number of corroded tiles than intact images with fewer 
(or no) tiles misclassified as corroded.

A value for the hyper-parameter $c$ can be chosen by optimizing a performance metric on the validation set.
More specifically, if $M(\left\{\hat{y}^{(i)}(c):i\right\})$ is a performance metric, viewed as a function of 
image predictions $\left\{\hat{y}^{(i)}(c):i\right\}$ dependent on $c$ and made over the validation set, then $c$ can be estimated 
by $\hat{c}=\sup_{c}M(\left\{\hat{y}^{(i)}(c):i\right\})$.

The F1 score is a plausible choice of performance metric $M(\left\{\hat{y}^{(i)}(c):i\right\})$ in the corrosion detection application
of this paper, since it places emphasis on true positives rather than true negatives and
since it can be applied on possibly imbalanced classes of intact and of corroded images
(see the Appendix for the definition of F1 score).

To summarize, the proposed algorithm for detecting whether an image contains corrosion comprises the 
following steps:
\begin{enumerate}
\item Run a ResNet on the validation set of tiles to tune hyperparameters relevant to training, such as the learning rate for 
stochastic gradient descent.
\item Train the ResNet on the training set of tiles.
\item Tune the value of hyperparameter $c$ using the validation set of tiles. To do so, predict the labels of tiles in the validation 
set via the trained ResNet, use these tile predictions to compute the whole image predictions via equation \eqref{eq:thres} for 
various values of $c$, and select the value of $c$ that maximizes a performance metric, such as the prediction accuracy or F1 
score.
\item Predict the labels of images in the test set. To do so, predict the labels of tiles in the test set via the trained ResNet, and 
use these tile predictions to compute the whole image predictions via equation \eqref{eq:thres} for the value of $c$ obtained at 
step 3.
\end{enumerate}

As it can be seen from the described algorithm, a new image is cropped into tiles, its tiles are predicted as corroded or intact via 
the trained ResNet, and equation \eqref{eq:thres} is used for predicting whether the image is corroded or intact.

In the regulatory guide of the
American Society of Mechanical Engineers (ASME) code cases not approved for use \cite{facco2017asme},
one metric for canister inspection is based on estimating the areal percent of the surface that is corroded.
The proposed algorithm in this paper offers a way of quantifying the percent of surface area that is corroded.
More specifically, the count $\sum_{t=1}^{n_i}\hat{y}_{t}^{(i)}$ of tiles predicted as corroded in equation \eqref{eq:thres}
divided by the total number of tiles per whole image
is a crude estimate of the percent of corroded surface area per whole image.

\subsection{Tuning, training and implementation}
\label{tuning_and_implementation}

This section provides details of hyperparameter tuning, of the configuration of the training process and of the implementation. 
To start with, data augmentation was performed on the training and validation sets
by carrying out random rotations, flips, color shifting, zooming and symmetric 
warping of the tiled images.

An $18$-layer, a $34$-layer, and a $50$-layer ResNet were employed in order to validate the discriminative power of the 
proposed approach to corrosion detection regardless of the deployed ResNet architecture (see
\cite{He_2016_CVPR}
for the definitions of ResNet-18, 
ResNet-34 and ResNet-50). 
The last fully connected layer of $1,000$ units in each of these three ResNets was replaced by
a fully connected layer of $2$ units.
From this point onwards, every reference to 
ResNet-18, ResNet-34 and ResNet-50 alludes to the respective ResNets with modified last layers.

\begin{figure*}[t!]
\centering
\subfloat[Loss vs learning rate;
ResNet-34, batch size 128]{{\includegraphics[scale=0.4]{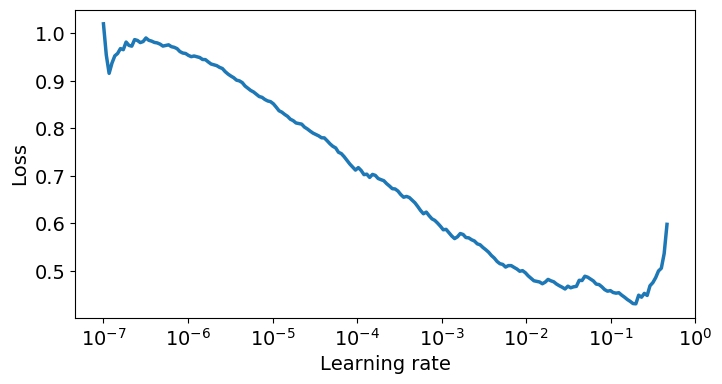}}
\label{fig:6a}}\qquad
\subfloat[F1 score vs $c$;
ResNet-34, batch size 128]{{\includegraphics[scale=0.4]{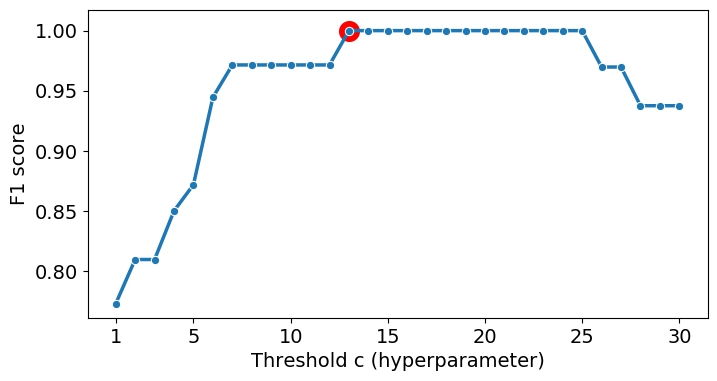}}
	\label{fig:6b}}\\
\subfloat[Loss vs epoch;
ResNet-34, batch size 32]{{\includegraphics[scale=0.4]{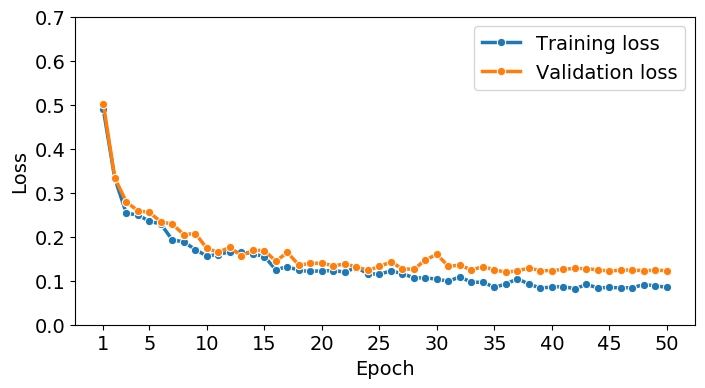}}
\label{fig:6c}}\qquad
\subfloat[Loss vs epoch;
ResNet-34, batch size 128]{{\includegraphics[scale=0.4]{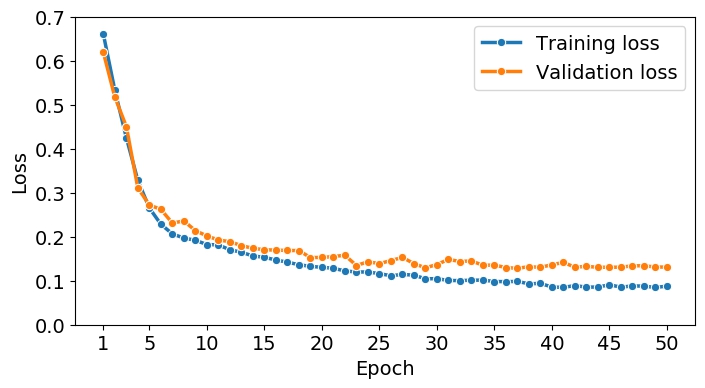}}
\label{fig:6d}}
\caption{
Diagnostic plots related to ResNet tuning and training.
(\ref{fig:6a}): plot of loss versus learning rate from a mock training session
with the learning rate finder of \cite{smith2015no}.
The training session was run using ResNet-34 and a batch size of $128$.
The loss has the sharpest downward slope in the vicinity of $5\cdot 10^{-4}$, so
the learning rate interval $[10^{-6},5\cdot 10^{-4}]$ is a plausible choice for discriminative layer training.
(\ref{fig:6b}): plot of F1 score versus hyperparameter $c$.
Tiled image predictions were made on the validation set via
ResNet-34 using a batch size of $128$.
The tiled image predictions
were then used for making whole image predictions via equation \eqref{eq:thres} for $c=1,2,\dots,30$.
The F1 score was computed for the image predictions arising from each value of $c$.
$\hat{c}=13$ is chosen as the threshold value 
maximizing the F1 score (see red-colored point).
(\ref{fig:6c}) and (\ref{fig:6d}):
plots of training and validation loss versus epoch obtained by training ResNet-34
with the one-cycle policy of \cite{smith2018disciplined} for batch sizes of $32$ and of $128$, respectively.
The two batch sizes yield similar training and validation loss curves.
}%
\label{fig:tuning}%
\end{figure*}

A mock training session with the learning rate finder of \cite{smith2015no} was run on the validation set
for each ResNet.
The learning rate was tuned across all layers, since the ResNet backbone was not kept frozen.
Figure \ref{fig:6a} displays an example of a loss curve
plotted against the range of learning rates used in a mock run
of the learning rate finder on ResNet-34.
According to Figure \ref{fig:6a}, the loss curve has a downward slope in the interval $[10^{-6},10^{-1}]$.
Following common practice, a value approximately in the middle of the sharpest downward slope,
such as $5\cdot 10^{-4}$, is chosen as the learning rate upper bound.

After tuning the learning rate,
discriminative layer training using the one-cycle policy of \cite{smith2018disciplined} was performed.
Discriminative layer training
refers to the process of training with different learning rates across layers.
On the basis of Figure \ref{fig:6a},
the learning rate range for discriminative layer training
was set to the interval $[10^{-6},5\cdot 10^{-4}]$.
This way, the learning rates for the first and last layer of each ResNet were set to $10^{-6}$ and to $5\cdot 10^{-4}$
during training, respectively,
with the intermediate layers having learning rates logarithmically distributed between $10^{-6}$ and $5\cdot 10^{-4}$.

Each ResNet was trained by running the one-cycle policy
\cite{smith2018disciplined}
for $50$ epochs.
Training was run for batch sizes of $32$, $64$, $128$ and $256$ for each of the three ResNets.
The training and validation loss curves were similar across different batch sizes and different ResNets.
For example, Figures \ref{fig:6c} and \ref{fig:6d} show a similar pattern of training and valitation loss curves for ResNet-34,
and for respective batch sizes of $32$ and $128$.

The threshold $c$ of equation \eqref{eq:thres} for image classification was selected by maximizing the F1 score of image 
predictions on the validation set (see Section \ref{results} for more details about the F1 score).
Figure \ref{fig:6b} provides an 
example of an F1 score curve, showing F1 values for a range of $c$ values.
It is noted that threshold $c$ is selected to maximize the F1 score
and does not translate to a corrosion threshold.
Whether an image is corroded or not is determined by acceptance criteria
mandated by acceptable industry standards.
However, different acceptance criteria will result in training sets
with different corroded/intact partitions
and subsequently in different values of threshold $c$.

A \texttt{Python} package, called \texttt{nccd}, accompanies this paper. It is publicly available at
\begin{center}
\texttt{https://github.com/papamarkou/nccd}
\end{center}
and it comes with an example of running the proposed algorithm. ResNet training and tile predictions in nccd make use of the 
\texttt{fastai} library. The \texttt{nccd} package implements the classification rule of equation \eqref{eq:thres} for making image 
predictions based on tile predictions. More generally, nccd automates the process of running the proposed algorithm. The EPRI 
data used in the analysis are proprietary, therefore they are not available in the public domain.

Simulations were run on an NVIDIA Tesla V100 GPU of a DGX-1 server at the CADES Cloud at Oak Ridge National Laboratory 
(ORNL). The training runtime for $50$ epochs using the one-cycle policy varied from $40$ minutes for ResNet-18 to $115$ 
minutes for ResNet-50.

\section{Results}
\label{results}

This section provides performance metrics on the basis of
tiled image predictions made by the three ResNets
and on the basis of whole image predictions made via equation \eqref{eq:thres}.
All predictions were made on the test set.
Tables \ref{table:2} and \ref{table:3} show numerical performance metrics, namely
true negatives (TN; intact correctly predicted as intact),
false positives (FP; intact wrongly predicted as corroded),
false negatives (FN; corroded wrongly predicted as intact),
true positives (TP; corroded correctly predicted as corroded),
true positive rates (TPR),
false positive rates (FPR),
positive predictive values (PPV) and
F1 scores.
The definition of these metrics are available in the Appendix.
Moreover, Figure \ref{fig:roc} displays
visual performance metrics,
namely receiver operating characteristic (ROC) curves,
which are plots of TPR against FPR.

\begin{table*}[t!]
	\centering
	\subfloat[Performance metrics for tiled images]{
\begin{tabular}{l|S[table-format=4.4]S[table-format=4.4]S[table-format=4.4]}
	\toprule 
	& \multicolumn{1}{c}{ResNet-18} & \multicolumn{1}{c}{ResNet-34} &  \multicolumn{1}{c}{ResNet-50} \\ \midrule
TN	& 7004	& 6936	& 6938 \\
FP	& 138	& 206	& 204 \\
FN	& 205	& 143	& 190 \\
TP	& 619	& 681	& 634 \\
TPR	& 0.7512	& 0.8265	& 0.7694 \\
FPR	& 0.0193	& 0.0288	& 0.0286 \\
PPV	& 0.8177	& 0.7678	& 0.7566 \\
F1	& 0.7830	& 0.7960	& 0.7629	 \\ \bottomrule
\end{tabular}
\label{table:2}
	} \qquad
	\subfloat[Performance metrics for whole images]{
	\begin{tabular}{l|S[table-format=2.4]S[table-format=2.4]S[table-format=2.4]}
	\toprule 
	& \multicolumn{1}{c}{ResNet-18} & \multicolumn{1}{c}{ResNet-34} &  \multicolumn{1}{c}{ResNet-50} \\ \midrule	
TN	& 16	& 15	& 14 \\
FP	& 0	& 1	& 2 \\
FN	& 1	& 1	& 0 \\
TP	& 16	& 16	& 17 \\
TPR	& 0.9412	& 0.9412	& 1 \\
FPR	& 0.0000	& 0.0625	& 0.1250 \\
PPV	& 1.0000	& 0.9412	& 0.8950 \\
F1	& 0.9697	& 0.9412	& 0.9444 \\ \bottomrule
\end{tabular}
\label{table:3}
	}
\caption{Performance metrics for assessing the quality of classification
	of tiled images (\ref{table:2}) and of whole images (\ref{table:3})
	as corroded or intact using three ResNets and a batch size of $128$.
	The first, second and third column in each table correspond to ResNet-18, ResNet-34 and 
	ResNet-50. TN, FP, FN, TP, TPR, FPR, PPV 
	and F1 stand for
	true negative, false positive, false negative, true positive, true 
	positive rate, false positive rate, positive predictive value 
	and F1 score, respectively. The performance 
	metrics were evaluated on the basis of tiled and whole image predictions made on the test set.}
	\label{table:2and3}
\end{table*}

\begin{figure*}[t!]
	\centering
	\subfloat[ROC curves;
	ResNet-34, four batch sizes]{{\includegraphics[scale=0.4]{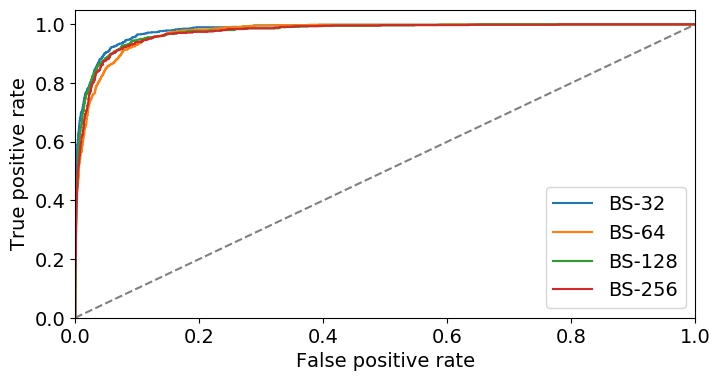}}
		\label{fig:7a}}\qquad
	\subfloat[ROC curves;
	batch size of $128$, three ResNets]{{\includegraphics[scale=0.4]{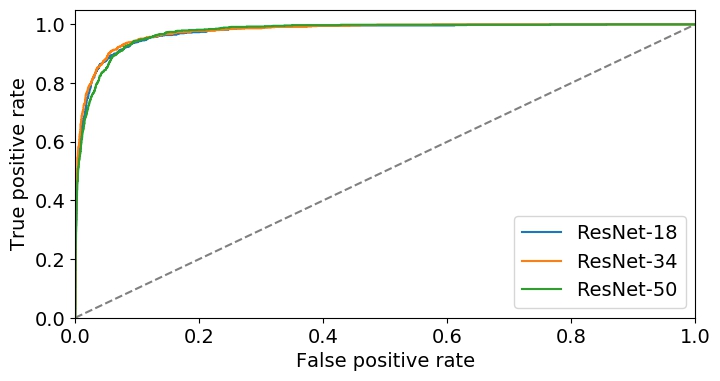}}
		\label{fig:7b}}
	\caption{
		ROC curves are nearly identical for different ResNets and for different batch sizes.
		All ROC curves were generated using the test set.
	}%
	\label{fig:roc}%
\end{figure*}

Tiled image predictions (on the test set) are similar across diffent batch sizes.
For example, the ROC curves for ResNet-34 based on four batch sizes ($32$, $64$, $128$ and $256$)
are nearly identical (Figure \ref{fig:7a}).
As another example, the ROC curves across the three ResNets for batch size equal to $128$
are nearly identical (Figure \ref{fig:7b}).
It is reminded that the training and validation loss curves (based on the training and validation set)
are also similar across different batch sizes (see Figures \ref{fig:6c} and \ref{fig:6d}).
In summary, neither the choice of ResNet nor the choice of batch size seem to have a drastic effect on
tiled image predictive capacity.
On the other hand,
a batch size of $128$ yields more accurate whole image predictions (on the EPRI test set)
than batch sizes of $32$, $64$ and $256$.
For this reason, the metrics of Tables \ref{table:2} and \ref{table:3}
pertain to predictions based on a batch size of $128$.

As explained in Section \ref{prop_alg},
the F1 score is preferred for tuning hyperparameter c in equation \eqref{eq:thres}.
As seen in Table \ref{table:2},
the F1 scores of tile predictions for ResNet-18, ResNet-34 and ResNet-50 are
$78.30\%$, $79.60\%$ and $76.29\%$.
The probabilities of corrosion detection (TPR) correspond to $75.12\%$, 
$82.65\%$ and $76.94\%$,
while the probabilities of false alarms (FPR) correspond to $1.93\%$, $2.88\%$ and $2.86\%$ for 
ResNet-18, ResNet-34 and ResNet-50.
Overall, tiled image predictive performance is similar across the three ResNet architectures
according to Table \ref{table:2}.

Moreover,
the image classification rule based on equation \eqref{eq:thres} is not sensitive to the choice 
of ResNet architecture according to Table \ref{table:3}.
More specifically and at whole image-level, ResNet-18, ResNet-34 
and ResNet-50 combined with the image classifier of equation \eqref{eq:thres} yield
F1 scores of $96.97\%$, $94.12\%$ and $94.44\%$,
probabilities of corrosion detection 
(TPR) of $94.12\%$, $94.12\%$ and $100\%$,
and probabilities of  false alarm (FPR) of $0\%$, $6.25\%$ and $12.50\%$.
Thereby, Table \ref{table:3} provides empirical evidence that the image classifier of equation \eqref{eq:thres} is 
not sensitive to tile misclassification errors or to the choice of ResNet architecture for tile classification.

So, this paper provides a first empirical indication that the algorithm of Section \ref{prop_alg}
can be used to automatically detect corrosion in the test images provided by EPRI.
At the same time,
it is emphasized that the EPRI test set consists of $33$ images only (see Table \ref{table:1a}).
A larger dataset would help to train, tune and test the proposed approach more extensively in the future.
Moreover, datasets collected from alternative sites
would help to assess the capacity of the proposed approach
to identify corrosion on spent fuel storage canisters with varying characteristics or from varying locations.

\begin{figure*}[t!]
\centering
\includegraphics[scale=0.175]{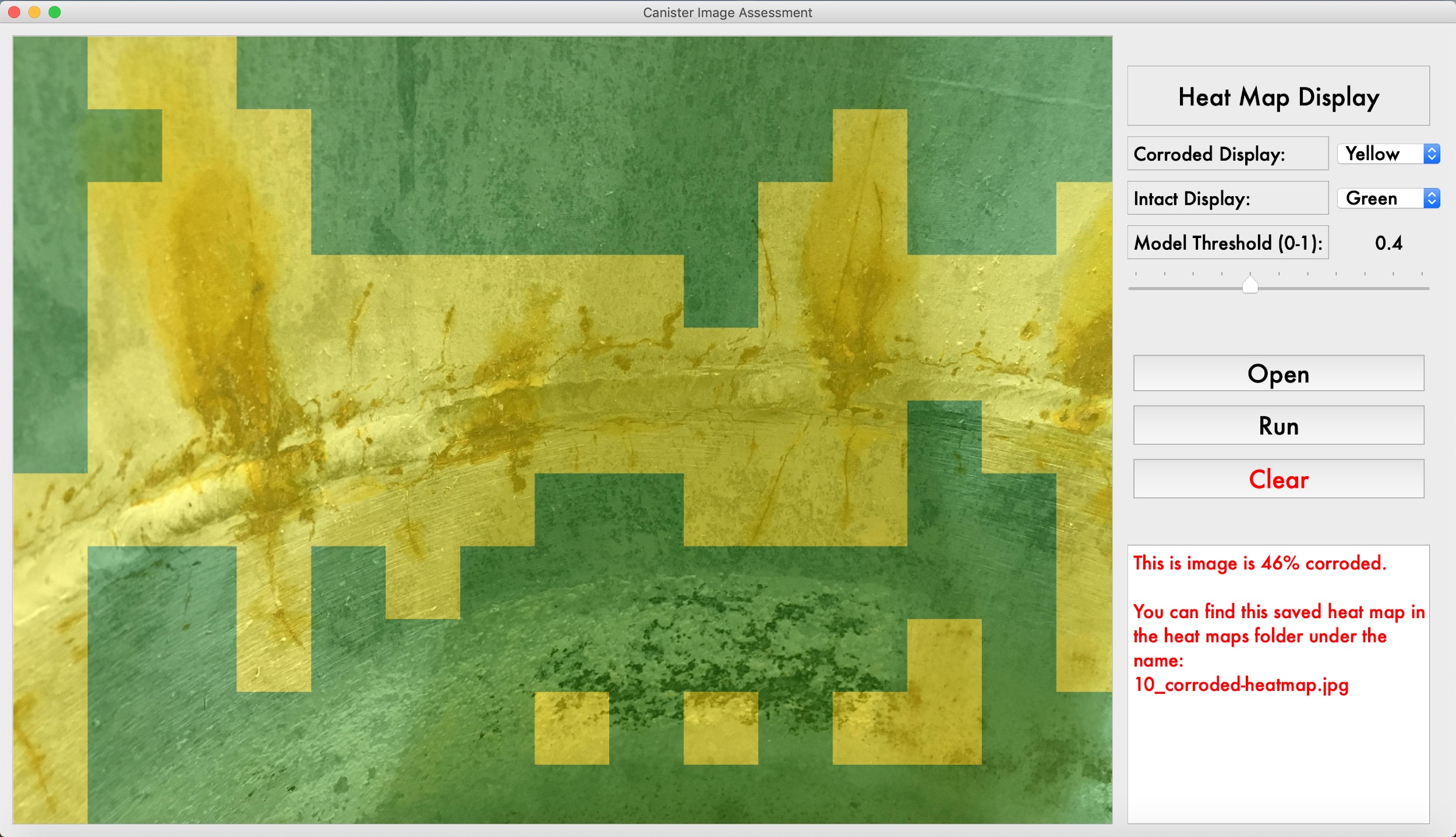}
\caption{Visualization of an image identified as corroded using the GUI. The corroded areas are highlighted using yellow color 
and intact areas using green color.}
\label{fig:gui}
\end{figure*}

A graphical user interface (GUI) was developed using \texttt{PyQt} to automate the process and analyze images. The GUI 
provides a colormap that highlights any identified corroded and intact areas within the image and provides additional information 
to the operator including percentage of image that is corroded. Figure \ref{fig:gui} demonstrates GUI usage via an image 
classified as corroded. It is noted that this image was particularly challenging with a lot of artifacts (shadows, scratches, etc.). 
However, the proposed deep learning approach captured correctly all the corroded areas (highlighted with yellow color). 

\section{Discussion and conclusions}

This paper proposes an image classification algorithm based on deep learning for detecting corrosion,
including discoloration, pitting, and stress corrosion cracks,
in stainless steel canisters. 
Despite the small sample size of $166$ EPRI images,
the algorithm attains high F1 score for whole images on the EPRI test set and 
therefore encourages further experimentation and investigation towards future deployment
as an operator decision support system intended to augment existing human 
inspector capabilities for detecting corrosion in used nuclear fuel canisters. The proposed algorithm trains a 
ResNet on a training set of tiles, tunes the value of hyperparameter $c$ using the validation set of tiles, and then uses tile 
predictions to compute whole image predictions.
The proposed image classifier is highly performing on the small EPRI test set of $33$ images and not sensitive to tile 
misclassification errors or to the choice of ResNet architecture for tile classification. Thus, the proposed algorithm holds promise 
for addressing the research question of automatically detecting corrosion in used fuel nuclear canisters.

Future work includes a more exhaustive testing by acquiring a larger dataset of images from a diverse collection of corroded and 
intact samples and by allowing for class imbalances at whole image-level. Variations of the classification rule of equation 
\eqref{eq:thres} can be considered, for example by introducing sliding window techniques across tiles to take into account 
spatial information in tile predictions along with tile prediction counts. Another future direction would be to attempt pixel-wise 
labelling instead of cropping into $256\times 256$ tiles, followed by a pixel-centric classification rule instead of the tile-centric 
classification rule of equation \eqref{eq:thres}. Such an approach would require of an upfront investment to annotate images 
pixel-wise. A preliminary attempt to assess the proposed approach using video input instead of image input has been made, 
which takes about 15 seconds for ResNet training per input file. To enable real-time usage with video input, future work will use 
optical flows. Video files are not efficient for solving the corrosion classification problem because consecutive frames contain 
redundant information as the camera pans. Optical flows eliminate this issue and allow for the entire canister to be visualized by 
a single image. Thus, the proposed algorithm will be evaluated on images derived from optical flows.


\appendix

\section{Performance metrics}

A tiled image prediction or an original image prediction
is characterized as positive if it is corroded (labeled as $1$), and as negative if it is intact 
(labeled as $0$). 
TPR is defined as
\begin{equation*}
TPR = TP/(TP+FN),
\end{equation*}
it provides the probability of detection of corroded images, and it is therefore crucial for assessing corrosion detection in nuclear 
canisters. FPR is defined as
\begin{equation*}
FPR = FP/(FP+TN),
\end{equation*}
and gives the probability of false alarm. PPV is defined as
\begin{equation*}
PPV = TP/(TP+FP),
\end{equation*}
and is particularly useful in the context of tile predictions, since the classes of intact and of corroded tiles are imbalanced with a 
sample size ratio of $37,762/4,957 \approx 7/1$ (see Table \ref{table:1b}). The F1 score is defined as
\begin{equation*}
F1 = 2\times PPV\times TPR/(PPV+TPR),
\end{equation*}
and it is the harmonic mean of FPR and PPV.
A ROC curve is a plot of TPR against FPR.

\section*{Acknowledgements}

This manuscript has been authored by UT-Battelle, LLC, under contract DE-AC05-00OR22725 with the US Department of Energy 
(DOE). The US government retains and the publisher, by accepting the article for publication, acknowledges that the US 
government retains a nonexclusive, paid-up, irrevocable, worldwide license to publish or reproduce the published form of this 
manuscript, or allow others to do so, for US government purposes. DOE will provide public access to these results of federally 
sponsored research in accordance with the DOE Public Access Plan (http://energy.gov/downloads/doe-public-access-plan).

This work was funded by the AI Initiative at the Oak Ridge National Laboratory.

This research used resources of the Compute and Data Environment for Science (CADES) at the Oak Ridge National Laboratory, 
which is supported by the Office of Science of the U.S. Department of Energy under Contract No. DE-AC05-00OR22725.

The authors would like to thank Guannan Zhang for helping with the co-supervision of the nuclear project interns at the artificial 
intelligence summer institute (AISI) 2019 of ORNL.




\begin{thebibliography}{10}
	\expandafter\ifx\csname url\endcsname\relax
	\def\url#1{\texttt{#1}}\fi
	\expandafter\ifx\csname urlprefix\endcsname\relax\def\urlprefix{URL }\fi
	\expandafter\ifx\csname href\endcsname\relax
	\def\href#1#2{#2} \def\path#1{#1}\fi
	
	\bibitem{kusnick2013finite}
	J.~Kusnick, M.~Benson, S.~Lyons, {Finite element analysis of weld residual
		stresses in austenitic stainless steel dry cask storage system canisters},
	Tech. rep. (2013).
	
	\bibitem{rigby2010evaluation}
	D.~B. Rigby, {Evaluation of the technical basis for extended dry storage and
		transportation of used nuclear fuel: executive summary}, US Nuclear Waste
	Technical Review Board, 2010.
	
	\bibitem{hanson2012gap}
	B.~Hanson, H.~Alsaed, C.~Stockman, D.~Enos, R.~Meyer, K.~Sorenson, {Gap
		analysis to support extended storage of used nuclear fuel (rev.0)}, Tech.
	rep., US Department of Energy Used Fuel Disposition Campaign (2012).
	
	\bibitem{chopra2014managing}
	O.~K. Chopra, D.~R. Diercks, R.~R. Fabian, Z.~H. Han, Y.~Y. Liu, {Managing
		aging effects on dry cask storage systems for extended long-term storage and
		transportation of used fuel (rev. 2)}, Tech. rep., Argonne National
	Laboratory, Argonne, IL (2014).
	
	\bibitem{chu2014flaw}
	S.~Chu, {Flaw growth and flaw tolerance assessment for dry cask storage
		canisters} (2014).
	
	\bibitem{Gorman2014}
	J.~Gorman, {Fuhr K.}, {Broussard J.}, {Literature Review of Environmental
		Conditions and Chloride-Induced Degradation Relevant to Stainless Steel
		Canisters in Dry Cask Storage Systems}, Tech. rep., Electric Power Research
	Institute, Palo Alto, CA (2014).
	
	\bibitem{FuhrK.2017}
	{Fuhr K.}, {Broussard J.}, {White G.}, {Aging management guidance to address
		potential chloride-induced stress corrosion cracking of welded stainless
		steel canisters}, Tech. rep., Electric Power Research Institute, Palo Alto,
	CA (2017).
	
	\bibitem{Broussard2015}
	J.~Broussard, S.~Chu, {Susceptibility assessment criteria for chloride-induced
		stress corrosion cracking (CISCC) of welded stainless steel canisters for dry
		cask storage systems}, Tech. rep., Electric Power Research Institute, Palo
	Alto, CA (2015).
	
	\bibitem{Meyer2016}
	R.~M. Meyer, {Nondestructive examination guidance for dry storage casks}, Tech.
	Rep. September, Pacific Northwest National Laboratory (2016).
	
	\bibitem{Meyer2013}
	R.~M. Meyer, {NDE to manage atmospheric SCC in canisters for dry storage of
		spent fuel}, Tech. rep., Pacific Northwest National Laboratory (2013).
	
	\bibitem{Renshaw2016b}
	J.~Renshaw, S.~Chu, {Dry canister storage system inspection and robotic
		delivery system development}, in: Transactions of the American Nuclear
	Society, Vol. 115, Palo Alto, CA, 2016, pp. 199--200.
	
	\bibitem{Chu2019}
	S.~Chu, {Dry cask storage welded stainless steel canister breach consequence
		analysis}, Tech. rep., Electric Power Research Institute, Palo Alto, CA
	(2019).
	
	\bibitem{BryanC.2016}
	{Bryan C.}, {Enos D.}, {Diablo Canyon Stainless Steel Dry Storage Canister
		Inspection}, Tech. rep., Electric Power Research Institute, Palo Alto, CA
	(2016).
	
	\bibitem{Tang2019}
	W.~Tang, S.~Chatzidakis, R.~Miller, J.~Chen, D.~Kyle, J.~Scaglione, C.~Schrad,
	{Welding process development for spent nuclear fuel canister repair}, in:
	American Society of Mechanical Engineers, Pressure Vessels and Piping
	Division (Publication) PVP, Vol.~1, 2019.
	\newblock \href {https://doi.org/10.1115/PVP2019-93946}
	{\path{doi:10.1115/PVP2019-93946}}.
	
	\bibitem{ChatzidakisStylianos2018}
	J.~M. {Chatzidakis, Stylianos and Adeniyi, Abiodun Idowu and Severynse, Tom F
		and Jones, Robert and Jarrell, Joshua and Scaglione}, {A novel mobile
		examination and remediation facility for On-site remediation of dry storage
		systems}, in: Proceedings of the Waste Management Conference, Phoenix, AZ,
	2018.
	
	\bibitem{StylianosChat2019}
	J.~J. {Stylianos Chatzidakis, Efe Kurt, Justin Coleman, John M Scaglione},
	{System-wide impacts of resolution options for a nonconforming dry storage
		system}, in: Proceedings of Global/TopFuel 2019, Seattle, Washington, 2019,
	pp. 1040--1047.
	
	\bibitem{Chatzidakis2020}
	S.~Chatzidakis, {Neutron diffraction illustrates residual stress behavior of
		welded alloys used as radioactive confinement boundary}, To appear (2020).
	
	\bibitem{Chatzidakis2019}
	S.~Chatzidakis, W.~Tang, J.~Chen, R.~Miller, A.~Payzant, J.~Bunn, J.~A. Wang,
	{Neutron residual stress mapping of repaired spent nuclear fuel welded
		stainless-steel canisters}, in: International High-Level Radioactive Waste
	Management 2019, IHLRWM 2019, 2019, pp. 249--253.
	
	\bibitem{Chatzidakis2018}
	S.~Chatzidakis, S.~Cetiner, H.~Santos-Villalobos, J.~J. Jarrell, J.~M.
	Scaglione, {Sensor requirements for detection and characterization of stress
		corrosion cracking in welded stainless steel canisters}, in: Proceedings of
	the 2018 International Congress on Advances in Nuclear Power Plants, ICAPP
	2018, 2018, pp. 714--720.
	
	\bibitem{Chatzidakis2017}
	S.~Chatzidakis, J.~J. Jarrell, J.~M. Scaglione, {High-resolution ultrasound
		imaging using model-based iterative reconstruction for canister degradation
		detection}, in: ANS IHLRWM 2017 - 16th International High-Level Radioactive
	Waste Management Conference: Creating a Safe and Secure Energy Future for
	Generations to Come - Driving Toward Long-Term Storage and Disposal, 2017,
	pp. 518--523.
	
	\bibitem{Lin2019}
	B.~Lin, D.~Dunn, R.~Meyer, {Evaluating the effectiveness of nondestructive
		examinations for spent fuel storage canisters}, in: AIP Conference
	Proceedings, Vol. 2102, 2019.
	\newblock \href {https://doi.org/10.1063/1.5099781}
	{\path{doi:10.1063/1.5099781}}.
	
	\bibitem{Renshaw2016a}
	{Dry canister storage system inspection and robotic delivery system
		development}, Tech. rep., Electric Power Research Institute, Palo Alto, CA
	(2016).
	
	\bibitem{Renshaw2015}
	J.~B. Renshaw, S.~Chu, J.~H. Kessler, K.~Waldrop, {Inspection and monitoring of
		dry canister storage systems}, in: 15th International High-Level Radioactive
	Waste Management Conference 2015, IHLRWM 2015, 2015, pp. 824--828.
	
	\bibitem{RenshawJ2019}
	M.~{Renshaw, Jeremy and Beard, Jamie and Stadler, Jim and Chu, Shannon and
		Orihuela}, {Robotically-deployed NDE inspection development for dry storage
		systems for used nuclear fuel}, in: International Conference on the
	Management of Spent Fuel from Nuclear Power Reactors: Learning from the Past,
	Enabling the Future, Vienna, Austria, 2019.
	
	\bibitem{ChuShannonand2019}
	A.~{Chu, Shannon and Renshaw, Jeremy and Akkurt, Hatice and Csontos}, {Aging
		management for dry storage canisters}, in: International Conference on the
	Management of Spent Fuel from Nuclear Power Reactors: Learning from the Past,
	Enabling the Future, Austria, Vienna, 2019.
	
	\bibitem{JonesJr.andRobert2015}
	H.~{Jones, Jr. and Robert}, {Dry storage cask inventory assessment. Prepared
		for DOE, nuclear fuels storage and transportation planning project}, Tech.
	rep., United States Department of Energy (2015).
	
	\bibitem{yeum2015vision}
	C.~M. Yeum, S.~J. Dyke, {Vision-based automated crack detection for bridge
		inspection}, Computer-Aided Civil and Infrastructure Engineering 30~(10)
	(2015) 759--770.
	
	\bibitem{rudin1992nonlinear}
	L.~I. Rudin, S.~Osher, E.~Fatemi, {Nonlinear total variation based noise
		removal algorithms}, Physica D: nonlinear phenomena 60~(1-4) (1992) 259--268.
	
	\bibitem{cha2016vision}
	Y.-J. Cha, K.~You, W.~Choi, {Vision-based detection of loosened bolts using the
		Hough transform and support vector machines}, Automation in Construction 71
	(2016) 181--188.
	
	\bibitem{o2014regionally}
	M.~O'Byrne, B.~Ghosh, F.~Schoefs, V.~Pakrashi, {Regionally enhanced multiphase
		segmentation technique for damaged surfaces}, Computer-Aided Civil and
	Infrastructure Engineering 29~(9) (2014) 644--658.
	
	\bibitem{wu2016improvement}
	L.~Wu, S.~Mokhtari, A.~Nazef, B.~Nam, H.-B. Yun, {Improvement of
		crack-detection accuracy using a novel crack defragmentation technique in
		image-based road assessment}, Journal of Computing in Civil Engineering
	30~(1) (2016) 4014118.
	
	\bibitem{lecun1998gradient}
	Y.~LeCun, L.~Bottou, Y.~Bengio, P.~Haffner, {Gradient-based learning applied to
		document recognition}, Proceedings of the IEEE 86~(11) (1998) 2278--2324.
	
	\bibitem{butcher2014defect}
	J.~B. Butcher, C.~R. Day, J.~C. Austin, P.~W. Haycock, D.~Verstraeten,
	B.~Schrauwen, {Defect detection in reinforced concrete using random neural
		architectures}, Computer-Aided Civil and Infrastructure Engineering 29~(3)
	(2014) 191--207.
	
	\bibitem{soukup2014convolutional}
	D.~Soukup, R.~Huber-M{\"{o}}rk, {Convolutional neural networks for steel
		surface defect detection from photometric stereo images}, in: International
	Symposium on Visual Computing, Springer, 2014, pp. 668--677.
	
	\bibitem{cha2017deep}
	Y.-J. Cha, W.~Choi, O.~B{\"{u}}y{\"{u}}k{\"{o}}zt{\"{u}}rk, {Deep
		learning-based crack damage detection using convolutional neural networks},
	Computer-Aided Civil and Infrastructure Engineering 32~(5) (2017) 361--378.
	
	\bibitem{chen2018}
	F.~Chen, M.~R. Jahanshahi, Nb-cnn: Deep learning-based crack detection using
	convolutional neural network and naïve bayes data fusion, IEEE Transactions
	on Industrial Electronics 65~(5) (2018) 4392--4400.
	
	\bibitem{mcculloch1943}
	W.~S. McCulloch, W.~Pitts, A logical calculus of the ideas immanent in nervous
	activity, The bulletin of mathematical biophysics 5~(4) (1943) 115--133.
	
	\bibitem{farley1954}
	B.~Farley, W.~Clark, Simulation of self-organizing systems by digital computer,
	Transactions of the IRE Professional Group on Information Theory 4~(4) (1954)
	76--84.
	
	\bibitem{rosenblatt1958}
	F.~Rosenblatt, The perceptron: a probabilistic model for information storage
	and organization in the brain., Psychological review 65~(6) (1958) 386.
	
	\bibitem{fukushima1980}
	K.~Fukushima, Neocognitron: A self-organizing neural network model for a
	mechanism of pattern recognition unaffected by shift in position, Biological
	cybernetics 36~(4) (1980) 193--202.
	
	\bibitem{lecun1989}
	Y.~LeCun, B.~Boser, J.~S. Denker, D.~Henderson, R.~E. Howard, W.~Hubbard, L.~D.
	Jackel, Backpropagation applied to handwritten zip code recognition, Neural
	computation 1~(4) (1989) 541--551.
	
	\bibitem{lecun1998}
	Y.~LeCun, L.~Bottou, Y.~Bengio, P.~Haffner, Gradient-based learning applied to
	document recognition, Proceedings of the IEEE 86~(11) (1998) 2278--2324.
	
	\bibitem{goodfellow2016deep}
	I.~Goodfellow, Y.~Bengio, A.~Courville, {Deep learning}, MIT press, 2016.
	
	\bibitem{cybenko1989}
	G.~Cybenko, Approximation by superpositions of a sigmoidal function,
	Mathematics of control, signals and systems 2~(4) (1989) 303--314.
	
	\bibitem{pratt1993}
	L.~Y. Pratt, Discriminability-based transfer between neural networks, in:
	Advances in neural information processing systems, 1993, pp. 204--211.
	
	\bibitem{He_2016_CVPR}
	K.~He, X.~Zhang, S.~Ren, J.~Sun, {Deep residual learning for image
		recognition}, in: The IEEE Conference on Computer Vision and Pattern
	Recognition (CVPR), 2016.
	
	\bibitem{facco2017asme}
	G.~Facco, {ASME} code cases not approved for use (rev.5), Tech. rep., Office of
	Nuclear Regulatory Research (2017).
	
	\bibitem{smith2015no}
	L.~N. Smith, {No more pesky learning rate guessing games}, CoRR, abs/1506.01186
	5 (2015).
	
	\bibitem{smith2018disciplined}
	L.~N. Smith, {A disciplined approach to neural network hyper-parameters: Part
		1--learning rate, batch size, momentum, and weight decay}, arXiv preprint
	arXiv:1803.09820 (2018).
	
\end{thebibliography}

\end{document}